\documentclass[10pt,twocolumn,letterpaper]{article}

\usepackage[pagenumbers]{cvpr} % To force page numbers, e.g. for an arXiv version
\usepackage{times}
\usepackage{epsfig}
\usepackage{graphicx}
\usepackage{color,soul}

\usepackage{caption}
\usepackage{subcaption}
\usepackage{svg}

\usepackage{amsmath,amsthm, amssymb}
\usepackage{mathrsfs}
\usepackage{mathtools}
\usepackage{microtype}
\usepackage{calligra}
\usepackage{amsfonts}
\usepackage{bbm}
\usepackage{yhmath}
\usepackage{bbold}
\usepackage{dsfont}
\usepackage{float}
\usepackage{multirow}
\usepackage{makecell}

\usepackage{enumerate}
\usepackage{xspace}
\usepackage{specialmacros}

\usepackage{tikz}
\usetikzlibrary{external,positioning,fit,calc,shapes}
\tikzset{set/.style={draw,circle,inner sep=0pt,align=center}}
\usepackage{tikzscale}
\usepackage{pgfplots}
\usepgfplotslibrary{groupplots,fillbetween, statistics}
\pgfplotsset{compat=newest}
\usetikzlibrary{math}
\usepackage{tabularx} % for 'tabularx' environment
\newcolumntype{P}[1]{>{\centering\arraybackslash}p{#1}}

\usepackage{dblfloatfix}

\newcommand\nnfootnote[1]{%
  \begin{NoHyper}
  \renewcommand\thefootnote{}\footnote{#1}%
  \addtocounter{footnote}{-1}%
  \end{NoHyper}
}

\defcitealias{kazhdan2013screened}{SPSR}
\defcitealias{huo2022iterative}{iPSR}
\defcitealias{hanocka2020point2mesh}{Point2Mesh}
\defcitealias{peng2021sap}{SAP}
\definecolor{cvprblue}{rgb}{0.21,0.49,0.74}
\usepackage[pagebackref,breaklinks,colorlinks,citecolor=cvprblue]{hyperref}
\colorlet{highlight}{cvprblue!30}
\sethlcolor{highlight}

\def\paperID{8572} % *** Enter the Paper ID here
\def\confName{CVPR}
\def\confYear{2024}

\def\ours{nPSR}

\title{Neural Poisson Surface Reconstruction: \\ Resolution-Agnostic Shape Reconstruction from Point Clouds}
\author{Hector Andrade-Loarca$^*$\\
Ludwig-Maximilians-Universität in  Munich\\
{\tt\small andrade@math.lmu.de}
\and
Julius Hege$^*$\\
Ludwig-Maximilians-Universität in Munich\\
{\tt\small hege@math.lmu.de}
\and
Daniel Cremers\\
Technical University of Munich\\
{\tt\small cremers@tum.de}
\and
Gitta Kutyniok\\
Ludwig-Maximilians-Universität in Munich\\
{\tt\small kutyniok@math.lmu.de}
}

\begin{document}
    \maketitle
    \iftoggle{cvprfinal}{\nnfootnote{$*$ Equal contribution.}}{}
    \begin{abstract}
We introduce Neural Poisson Surface Reconstruction (\ours{}), an architecture for shape reconstruction that addresses the challenge of recovering 3D shapes from points. Traditional deep neural networks face challenges with common 3D shape discretization techniques due to their computational complexity at higher resolutions. To overcome this, we leverage Fourier Neural Operators to solve the Poisson equation and reconstruct a mesh from oriented point cloud measurements. \ours{} exhibits two main advantages: First, it enables efficient training on low-resolution data while achieving comparable performance at high-resolution evaluation, thanks to the resolution-agnostic nature of FNOs. This feature allows for one-shot super-resolution. Second, our method surpasses existing approaches in reconstruction quality while being differentiable and robust with respect to point sampling rates. Overall, the neural Poisson surface reconstruction not only improves upon the limitations of classical deep neural networks in shape reconstruction but also achieves superior results in terms of reconstruction quality, running time, and resolution agnosticism.
\end{abstract}    
    \section{Introduction}
Shape reconstruction, the process of creating a three-dimensional mesh of an object from 2D images or point clouds plays a crucial role in computer vision. It is an active research area with significant applications in robotics, virtual reality, and autonomous driving. A common subproblem in this field is the task of reconstructing a surface from noisy point clouds, which are typically acquired using techniques such as photogrammetry (multiple images) or depth sensors.

A plethora of shape reconstruction algorithms have been developed, as surveyed in \cite{berger2017survey}. Recently, deep learning approaches trained on large datasets of 3D objects, such as Convolutional Occupancy Networks \cite{peng2020convolutional} and POCO \cite{boulch2022poco}, have emerged, promising improved reconstruction quality. However, these methods typically do not generalize well beyond the training distribution \cite{sulzer2023dsr}.

Another challenge faced by deep learning approaches is scalability, especially when dealing with complex models at high resolutions. Convolutional neural networks have achieved widespread success in computer vision and can straightforwardly generalize to three dimensions. However, they operate on regular grids which limits their ability to represent fine details accurately.

To overcome this limitation, various strategies have been proposed. Some solutions involve working with irregular grids \cite{peng2020convolutional}, representing shapes as implicit functions  \cite{chibane2020implicit}, or employing multi-grid approaches in a coarse-to-fine manner \cite{sharf2006coarse2fine} or sliding-window fashion \cite{nguyen2021slidingwindow}.

Our aim in this work is to enhance reconstruction quality by integrating deep learning into the well-established Poisson surface reconstruction method, thereby leveraging robustness and generalization capabilities. In order to achieve this, we introduce a neural network architecture whose backbone is a \emph{Fourier Neural Operator} \cite{li2021fourier}. The choice is motivated by Fourier Neural Operators' ability to approximate partial differential equation solutions.

\subsection{Poisson Surface Reconstruction}
\label{sec:Poisson}
We begin by introducing the problem of Poisson surface reconstruction. In this setting, we are given a point cloud sampled from the surface of a (three-dimensional) object and the corresponding normal vectors, and we want to recover the indicator function that implicitly represents the shape of such an object. Assuming that we have access to the full normal vector field $V:\Omega\subset \mathbbm{R}^3 \rightarrow {\mathbbm{R}^3}$, our goal is to reconstruct/learn a function $\chi:\Omega \rightarrow \lbrace 0,1\rbrace$ such that
\begin{align} \label{eq:normal}
    \nabla \chi(x) = V(x) \quad \text{for all }x\in \Omega,
\end{align} where $\chi$ denotes the indicator function representing the object. 
Taking the divergence of \eqref{eq:normal}, we obtain the Poisson equation
\begin{align} \label{eq:poisson}
    \Delta \chi(x) = \nabla \cdot V(x) \quad \text{for all }x\in \Omega.
\end{align} 
Notice that the indicator function is unique up to a constant. The uniqueness of a solution can be ensured by imposing sparse constraints. Here, the domain $\Omega$ can be chosen to be Lipschitz and large enough so that the object is fully contained in $\Omega$, i.e., $\chi(x)=0$ on $\partial \Omega$. Thus, we face a boundary value problem given by \eqref{eq:poisson} supplemented with homogeneous Dirichlet boundary conditions. This is an ill-posed inverse problem since for any given finite point cloud there exists multiple possible solutions.

\subsection{Fourier Neural Operator} \label{se:FNO}

Fourier Neural Operators (FNOs) were first introduced by Li et al. in \cite{li2021fourier} as a technique to learn the solution of PDEs in the form of operators. In general terms, the aim of an FNO is to learn a mapping $\mathcal{N}:\mathcal{A}\rightarrow \mathcal{U}$ between two infinite-dimensional function spaces by using a finite collection of observations $\lbrace (a_i,u_i)\rbrace_{i=1}^N$, where $\mathcal{N}(a_i)=u_i$ for all $i$.

\begin{definition}[Fourier Neural Operator, \cite{li2021fourier}]
\label{def:fno}
A \emph{Fourier Neural Operator} is given by the iterative architecture
\begin{equation}
\label{eq:fno}
\begin{aligned}
a\rightarrow P a &=:v_0\mapsto \cdots \mapsto v_T \mapsto Q v_T =: u,\\
v_{t+1}(x)&:=\sigma(Wv_t(x)+\mathcal{F}^{-1}\left(R \cdot(\mathcal{F}v_t)\right))(x)
\end{aligned}
\end{equation}
for $t=0, \dots, T-1,$ where $\mathcal{F}$ denotes the Fourier transform, $R\in \mathbb{C}^{k_\text{max}{\times}d_v{\times}d_v}$ represents the weight tensor, ${P:\mathcal{A} \rightarrow \mathbb{R}^{d_v}}$ is a \emph{lifting operator}, $Q:\mathbb{R}^{d_v}\rightarrow \mathcal{U}$ is a \emph{projection operator}, $W:\mathbb{R}^{d_v}\rightarrow \mathbb{R}^{d_v}$ is a linear transformation, and ${\sigma:\mathbb{R}\rightarrow \mathbb{R}}$ is a nonlinear activation function applied component-wise.
\end{definition}
\noindent \cref{fig:NeuralPoisson} shows the schematic representation of the architecture of a Fourier Neural Operator (FNO) for four Fourier blocks. The aim of the FNO is to learn the weights $R$. One can interpret $R$ as the Fourier representation of a convolutional kernel, where the formula $\mathcal{F}^{-1}
\left(R\cdot(\mathcal{F}v_t)\right)$ in~\eqref{eq:fno} is obtained by the \emph{Fourier convolution theorem}. Now, in order to implement this method, we need to discretize the layer~\eqref{eq:fno}.

%\vspace{-1em}

\subsection{Discretization}
We assume that our domain $\Omega\subset \Real^3$, where the input function $a\in \mathcal{A}$ is evaluated at, is discretized into $d_a\in \mathbb{N}$ points and we wish to evaluate the output function at a higher resolution with $d_u \in \mathbb{N}$ points. Thus, the lifting and projection operator can be represented by the matrices $P\in \mathbb{R}^{d_v{\times}d_a}$ and $Q\in \mathbb{R}^{d_u{\times}d_v}$, respectively. 
Furthermore, in~\eqref{eq:fno} we have, $v_t\in \Real^{d_a{\times}d_v}$ and $\mathcal{F}(v_t)\in \mathbb{C}^{d_a{\times}d_v}$. The key advantage of Fourier layers (over traditional convnets) is that they are discretization-invariant since parameters are learned directly in Fourier space. Resolving the functions in physical space simply amounts to projecting on the basis of wave functions which are well-defined everywhere in the space. This allows us to transfer the learned network among resolutions. If implemented with standard FFT, then it will be restricted to uniform mesh, but still independent of discretization, therefore \emph{resolution-agnostic}. This feature of the Fourier Neural Operator is crucial in the context of shape since 3D data requires a lot of memory. 

In this work, we also assume the discretization to be uniform, thus $\mathcal{F}$ can be replaced by the Fast Fourier Transform (FFT). 
Furthermore, following \cite{li2021fourier}, we truncate the higher Fourier modes of the FFT by keeping the first $k_\text{max}$ Fourier modes. The multiplication by the weight tensor $R\in \mathbb{C}^{k_\text{max}{\times}d_v{\times}d_v}$ is then given by

\begin{equation}
\begin{aligned}
 \left(R\cdot (\mathcal{F}v_t)\right)_{k,l}=\sum_{j=1}^{d_v}R_{k,l,j}(\mathcal{F}v_t)_{k,j}, 
 %\vspace{-2em}
\end{aligned}  
\end{equation}
$k=1,\ldots, k_\text{max}, j=1,\ldots, k_\text{max}.$

\section{Related Work}

\subsection{Poisson Surface Reconstruction}

The use of the Poisson equation to solve a 3D shape reconstruction problem was first introduced by Kazhdan et al. in 2006 \cite{kazhdan2006poisson}. Later, the same authors proposed an improvement by adding additional boundary conditions named \emph{screened Poisson surface reconstruction (SPSR)} \cite{kazhdan2013screened}. 

More recently, Poisson surface reconstruction has been further explored, first in the form of a differentiable Poisson solver (Peng et al. \cite{peng2021sap}) which can be used to either optimize a Poisson surface reconstruction or learn it. In addition, Hue et al. introduced in 2022 an iterative method that can perform the reconstruction without the need of normals, the \emph{iterative Poisson surface reconstruction (iPSR)} \cite{huo2022iterative}. Their method initializes the normals of a given point cloud randomly and denoises them by iteratively applying the Poisson surface reconstruction. Sell{\'a}n et al. \cite{sellan2023neural} have recently proposed solving a stochastic form of the Poisson equation using Physics-Informed Neural Networks, optimizing a neural network to represent an object. 

These techniques have shown amazing performance in recovering high-resolution close-to-perfect reconstructions in high-sampling regimes ($25\,0000$ points). Yet, in the low sampling regime (around $3\,000$ to $25\,000$ points) they often generate very sub-optimal reconstructions, lacking the ability to correctly interpolate surfaces. Furthermore, their computational expense is significantly influenced by the input resolution, making them inefficient for high-resolution applications.

\subsection{Fourier Neural Operators}

In the past years, \emph{Fourier Neural Operators (FNOs)} have emerged as a potent tool for differentiably solving complex, high-dimensional PDEs \cite{li2021fourier}, thanks to their training convenience and resolution-agnostic attributes.  Given these advantages, FNOs have been employed in tasks demanding high resolution, pertaining to traditional PDEs, such as weather modeling \cite{pathak2022fnoweather} and multiphase flow \cite{wen2022multiphase}. 

Motivated by the poor performance of existing methods in the low-sampling regime, we introduce an approach that addresses this drawback by using the ability of FNOs to solve PDEs in challenging scenarios such as low-sampling and high-resolutions. 

\section{Main Contribution}

We present a new approach for solving the problem of 3D shape Poisson surface reconstruction. Our method outperforms existing methods, particularly in low sampling scenarios where the input point cloud consists of approximately $3\,000$ to $25\,000$ points. In addition, in high-sampling regimes, it performs as well as the classical method which presents close-to-perfect reconstruction given sufficiently many points (around $25\,0000$ points in our case).

Our approach is based on a deep neural network architecture formed by Fourier blocks and can thus be considered a Fourier Neural Operator, which is used to solve the underlying Poisson equation. It takes a directed point cloud as input and outputs a 3D reconstruction in the form of a voxel grid. We further threshold the output using Otsu's method \cite{otsu1979threshold} and finally perform marching cubes \cite{lorensen1987marching} to extract a mesh. 

In addition to outperforming existing approaches, our method also inherits the one-shot super-resolution capabilities of Fourier Neural Operators. We are able to train the method on voxel grids of size $64{\times}64{\times}64$ while evaluating it on grids of size $128{\times}128{\times}128$ with similar performance, in contrast to classical convolutional neural networks where the error grows exponentially with the resolution of the input. In essence, our technique consistently produces superior-quality reconstructions and demonstrates enhanced robustness when considering variations in point sampling rate and resolution, surpassing the performance of current methods.

Finally, notably, this study pioneers the application of a Fourier Neural Operator in computer vision tasks, highlighting its prospective significance in this field.

    \section{Method}

In this section, we will explain in detail all the steps that our method follows to learn how to compute a watertight mesh from a given oriented point cloud. 

\subsection{Preparation of the Training Data}
\label{subsec:PrepData}

To train our model, we utilize the ShapeNet dataset \cite{chang2015shapenet} (see \url{https://shapenet.org/}). Since the standard output of a Poisson surface reconstruction algorithm is a voxel grid of a given dimension, we use the voxelized meshes provided in ShapeNet to obtain clean ground truths. Additionally, we generate ground truth data for the oriented point cloud samples by performing marching cubes \cite{lorensen1987marching} on the voxel grids and sampling points on the resulting surface mesh using the Poisson disk sampling technique proposed by Yuksel in \cite{yuksel2015disksampl}.

Since we extract the meshes using marching cubes, we have access to the normals of their faces. All sampled points are on some triangle of the ground truth mesh. We calculate the normals of the sampled points using the average over the neighboring faces, weighted by their barycentric coordinates.

We trained our method using four different sampling scenarios, with $3\,000$, $10\,000$, $25\,000$, and $250\,000$ sampling points, all of them with resolution $64\times 64\times 64$. These experiments demonstrate how our method outperforms other methods in the low sampling regime ($3\,000$-$25\,000$) while producing close-to-perfect reconstructions in the high sampling regime ($250\,000$), comparable to other methods, including the classical Poisson surface reconstruction \cite{kazhdan2006poisson}. In addition, we also demonstrate resolution-agnosticism by comparing the reconstruction quality when the resolution is changed. One can find the numerical results and reconstruction images corresponding to these experiments in~\cref{sec:experiments}.

\subsection{Architecture and Training Procedure}

\begin{figure}[htb!]
      \centering
    \includegraphics[width=\linewidth]{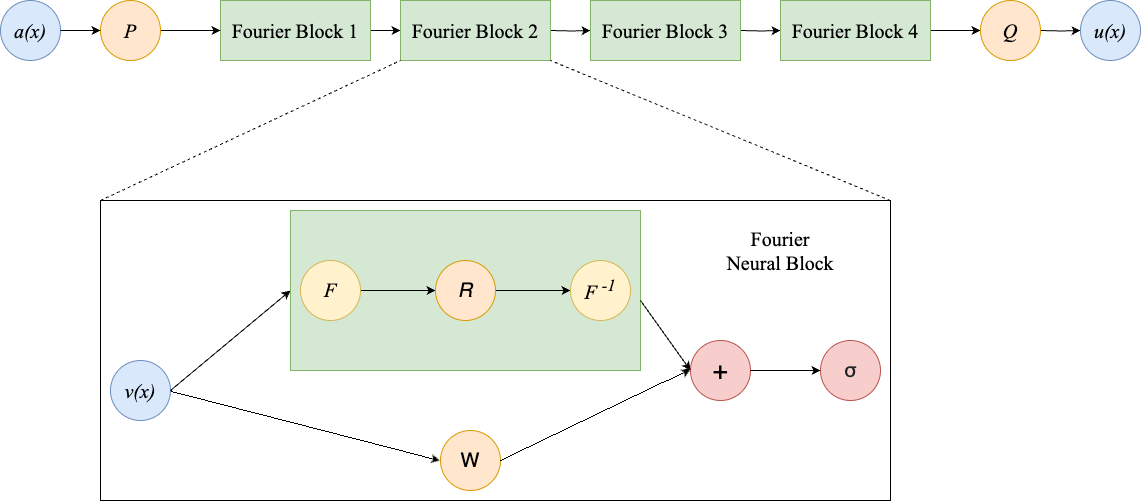}
\caption{Fourier Neural Operator with four Fourier Blocks, the backbone of \ours{}, where $\mathcal{F}$ represents the Fourier transform.}
\label{fig:NeuralPoisson}
\end{figure}

The full pipeline of our approach is composed of a Fourier Neural Operator with four 3D Fourier blocks (see \cref{fig:NeuralPoisson}) as well as a pre-processing and a post-processing step. Following each Fourier block, we incorporate GELU (Gaussian Error Linear Unit, \cite{hendrycks2016gaussian}) non-linearities and dropout layers after the first three Fourier blocks. Furthermore, we employ $20$ frequency modes on the Fourier transform.

\ours{} is trained on a representative subset of $10\,000$ randomly selected shapes from the ShapeNet dataset, which accounts for a fifth of the full dataset.The preparation of the data is detailed in \cref{subsec:PrepData}, outlining our comprehensive approach. The complete pipeline is visualized in \cref{fig:Pipeline}.

To assess our model's performance, we use the mean squared error on the voxel grid as the loss function, consistent with prior works on 3D FNOs \cite{li2021fourier,wen2022ufno}. Furthermore, we evaluate and test the network on two distinct sets, each containing $2\,000$ examples. Throughout training, we employ batches of 20 data points for approximately $10\,000$ training steps.

\subsection{Data pre-Processing and Prediction post-Processing}
\label{subsec:DataPrep}

As discussed in the previous sections, the input of our method is regarded as a directed point cloud, consisting of point coordinates and normal directions associated with each point. The Fourier block (see \cref{fig:NeuralPoisson}) we use to construct our Fourier Neural Operator architecture, \ours{}, assumes the input to be defined in a uniform grid. In order to transform our input point cloud into a uniform grid representation we perform point rasterization (see \cite{noll1971raster}) of the divergence field $\nabla\cdot V$ in \eqref{eq:poisson}. The point rasterization is done in two steps, we first rasterize the point locations in a voxel grid of a given resolution (e.g. $64\times 64\times 64$), where we perform a smoothing step using a Gaussian filter of standard deviation $\sigma = 2$. The second step consists of incorporating the normal information of every given point to compute the divergence field with finite differences. Finally, we perform a smoothing step on the resulting divergence field using the filter from the first step. The resulting voxel grid representing the divergence field is then fed into our Fourier neural architecture. 

\begin{figure*}[htb!]
      \centering
    \includegraphics[width=\linewidth]{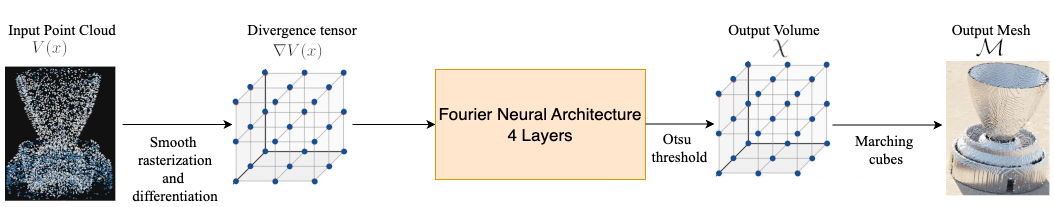}
\caption{\ours{}, full pipeline.}
\label{fig:Pipeline}
\end{figure*}

To refine \ours{}'s reconstruction, we employ a thresholding step as a post-processing technique. The threshold value is computed through Otsu's thresholding method \cite{otsu1979threshold}. Finally, we generate a surface mesh through the marching cubes algorithm \cite{lorensen1987marching}. The complete pipeline is illustrated in \cref{fig:Pipeline}. 

    \section{Experiments}
\label{sec:experiments}

In this section, we present the numerical results as well as some reconstruction examples. In addition, we also perform a number of ablation studies, which are presented in this section as well. All code and instructions necessary to reproduce our experiments can be found in the supplementary material.

\begin{table*}[bht]
\centering
\begin{tabular}{ cccccc} 

    \toprule
    
    N. Sampled Points & Method & Chamfer-L1 ($\downarrow$) & F-Score ($\uparrow$) & Norm. Consist. ($\uparrow$) & Runtime (s)\\ 
    
    \midrule
    
    \multirow{5}{*}{$3\, 000$}& SPSR \cite{kazhdan2013screened} & $0.430$ & $0.552$ & $0.525$ & \hl{$0.31$} \\ 
    & iPSR \cite{huo2022iterative} & $0.554$ & $0.488$ & $0.465$ & $325.25$ \\ 
    & Point2Mesh \cite{hanocka2020point2mesh} & $0.644$ & $0.392$ & $0.381$ & $2\,215.64$\\ 
    & SAP \cite{peng2021sap} & $0.492$ & $0.516$ & $0.495$ & $0.45$\\

    \cmidrule{2-6}
    
    & \textbf{\ours{}} & \hl{$0.081$} & \hl{$0.895$} & \hl{$0.875$} & $0.52$\\  
    
    \midrule
    
    \multirow{ 5}{*}{$10\,000$} & SPSR \cite{kazhdan2013screened} & $0.265$ & $0.722$ & $0.715$ & \hl{$0.31$} \\ 
    & iPSR \cite{huo2022iterative} & $0.372$ & $0.548$ & $0.625$ & $410.41$ \\ 
    & Point2Mesh \cite{hanocka2020point2mesh} & $0.485$ & $0.522$  & $0.578$ & $3\,015.91$\\ 
    & SAP \cite{peng2021sap} & $0.338$ & $0.852$ & $0.681$ & $0.45$\\
    
    \cmidrule{2-6}
    
    & \textbf{\ours{}} & \hl{$0.042$} & \hl{$0.954$} & \hl{$0.912$} & $0.52$ \\ 
    
    \midrule
    
    \multirow{ 5}{*}{$25\,000$} & SPSR \cite{kazhdan2013screened} & $0.230$ & $0.751$ & $0.753$ & \hl{$0.31$} \\ 
    & iPSR \cite{huo2022iterative} & $0.348$ & $0.583$ & $0.688$ & $452.33$\\ 
    & Point2Mesh \cite{hanocka2020point2mesh} & $0.412$ & $0.512$ & $0.623$ & $3\,890.82$\\ 
    & SAP \cite{peng2021sap} & $0.315$ & $0.88$ & $0.715$ & $0.45$\\

    \cmidrule{2-6}
    
    & \textbf{\ours{}} & \hl{$0.029$} & \hl{$0.983$} & \hl{$0.976$} & $0.52$\\ 
    
    \midrule
    
    \multirow{ 5}{*}{$250\,000$} & SPSR \cite{kazhdan2013screened} & $0.022$ & $0.954$ & $0.925$ & \hl{$0.31$} \\ 
     & iPSR \cite{huo2022iterative} & $0.031$ & $0.961$ & $0.930$ & $11\,015.84$ \\ 
     & Point2Mesh \cite{hanocka2020point2mesh} & $0.649$ & $0.614$ & $0.622$ & $43\,850.11$ \\ 
     & SAP \cite{peng2021sap} & $0.015$ & $0.974$ & $0.971$ & $0.45$ \\

    \cmidrule{2-6}
     
     & \textbf{\ours{}} & \hl{$0.008$} & \hl{$0.99$} & \hl{$0.991$} & $0.52$ \\ 
    
    \bottomrule
 
 \end{tabular}

\caption{Reconstruction metrics averaged over all ShapeNet voxel models. Comparisons against the baseline models SPSR \cite{kazhdan2013screened}, iPSR \cite{huo2022iterative}, Point2Mesh \cite{hanocka2020point2mesh} and SAP \cite{peng2021sap}.  For each metric, the best result is \hl{highlighted}. \ours{} achieves the best performance with a marginal increase in runtime. The performance improvement is around an order of magnitude in terms of Chamfer-L$1$ distance.}

\label{table:reconstruction_metrics}
\end{table*}

\begin{table*}[htb!]
\centering
\begin{tabular}{cccc} 

    \toprule
    
    Resolution (voxels) & Chamfer-L1 ($\downarrow$) & F-Score ($\uparrow$) & Norm. Consist. ($\uparrow$)\\ 
    
    \midrule

     $32\times 32\times 32$ & 0.024 & 0.992 & 0.983 \\
    
    \midrule
    
     $64\times 64\times 64$ & 0.027 & 0.986 & 0.980 \\
    
    \midrule
    
    $128\times 128\times 128$ & 0.029 & 0.983 & 0.976 \\

    \bottomrule
 
 \end{tabular}

\caption{Reconstruction metrics averaged over all ShapeNet voxel models, with $25\,000$ sampled points. Comparison of reconstruction performance along different resolutions}

\label{table:resolution_agnostic}
\end{table*}

\begin{figure*}[b!]
    \centering
    \begin{subfigure}[b]{.725\linewidth}
        \centering
        \includegraphics{figs/table.tikz}
        % \caption{}
        % \label{fig:shapenet-3}
    \end{subfigure}%
\end{figure*}
\begin{figure*}[t!]
    \ContinuedFloat
    \centering
    \begin{subfigure}{.725\linewidth}
        \centering
        \includegraphics{figs/basket.tikz}
        % \caption{}
        % \label{fig:shapenet-1}
    \end{subfigure}
\end{figure*}
\begin{figure*}
    \ContinuedFloat
    \centering
    %\hfill
    \vspace{-.94\baselineskip}
    \begin{subfigure}[b]{.725\linewidth}
        \centering
        \includegraphics{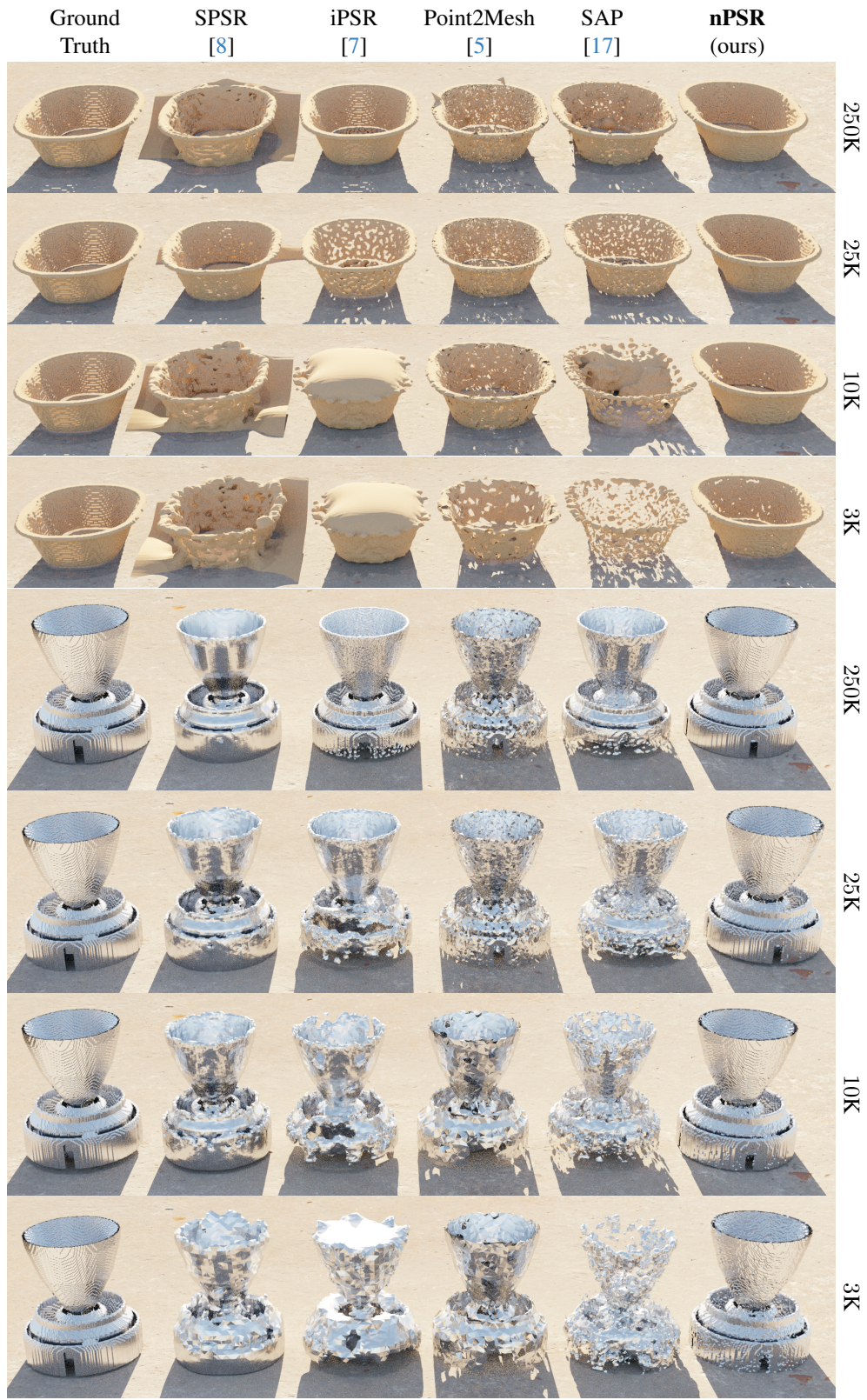}
        % \caption{}
        % \label{fig:shapenet-2}
    \end{subfigure}%
    \caption{While all methods work well in high-sampling regimes, the proposed \ours{} significantly outperforms in low-sampling regimes.}
    \label{fig:shapenet}
\end{figure*}

\begin{figure*}
    \centering
    \begin{subfigure}[b]{.725\linewidth}
        \centering
        \includegraphics{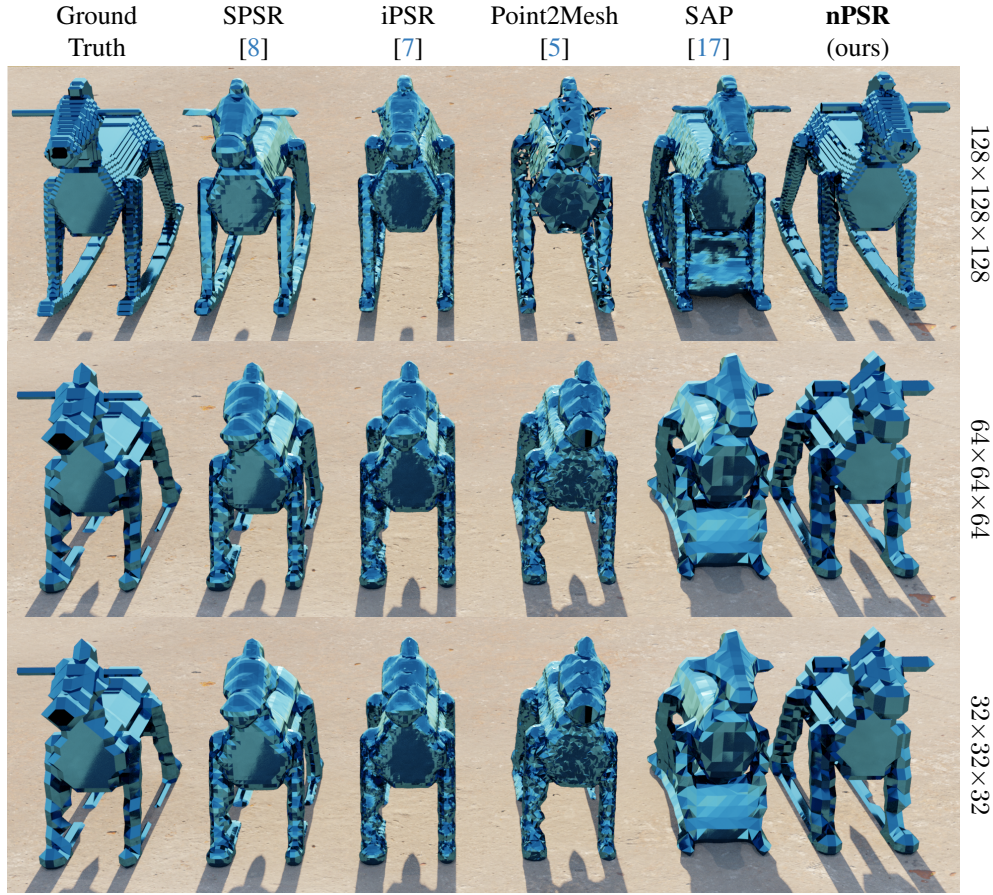}
     \end{subfigure}
\caption{Our method is resolution agnostic, performing much better than existing methods when the resolution increases in inference time.}
\label{fig:multires}
\end{figure*}

\subsection{Results}
We conduct experiments with four different sampling rates, $3\,000$, $10\,000$, $25\,000$, and $250\,000$ sampled points (and normals). In addition, the performance of our method is also compared on the same data with other four existing methods, namely, Screened Poisson Surface Reconstruction (SPSR) \cite{kazhdan2013screened}, Iterative Poisson Surface Reconstruction (iPSR) \cite{huo2022iterative}, Point2Mesh \cite{hanocka2020point2mesh} and Shape as Points (SAP) \cite{peng2021sap}. To assess the quality of reconstructions, we utilized metrics introduced by Peng et al. \cite{peng2021sap}, which include Chamfer L1 loss (measuring point cloud distances), F-Score (evaluating point estimation classification accuracy), and Normal Consistency (accounting for normal orientation). All experiments were performed on the ShapeNet Core dataset (\url{https://shapenet.org/}).

Our method was trained for approximately $1\,000$ epochs on shapes with resolution $64{\times}64{\times}64$ and evaluated on shapes with resolution $128{\times}128{\times}128$. The numerical results for the different sampling rates are presented in \cref{table:reconstruction_metrics}. Notably, \ours{} consistently outperformed the alternatives in all sampling rates, with particularly significant improvements observed in the very low sampling regime ($3\,000$). In contrast, other methods exhibited poorer performance in such scenarios.

While our approach may not be the fastest (classical SPSR holds that distinction), it maintains a constant running time relative to the sampling rate. This is because most of the point cloud processing, such as rasterization, occurs during pre-processing at the training stage, enabling straightforward scalability.

We also perform reconstructions for different resolutions, in which the model was trained on resolution $64{\times}64{\times}64$ with $25\,000$ sampled points and evaluated on resolutions $32{\times}32{\times}32$ and $128{\times}128{\times}128$. \cref{{table:resolution_agnostic}} depicts the performance metrics corresponding to the different resolutions, showing the robustness of the method against different resolutions (resolution-agnosticism). 

Finally, \cref{fig:shapenet} showcases three examples of reconstructions at different sampling rates, visually confirming the superior performance of our method over the others. In addition, \cref{fig:multires} shows a reconstruction example with different resolutions, confirming the resolution-agnosticism of the method and the superiority with respect to other methods.

\subsection{Ablation Studies}
We conducted various ablation studies to optimize our pipeline and model setup. Initially, the input was computed without incorporating the normal information, relying solely on rasterization and smoothing of the point cloud, which led to unsatisfactory performance. This highlighted the crucial role of normal field information in our model. Subsequently, we experimented with different parameters for the smoothing Gaussian used in the input, ultimately discovering that a standard deviation of $\sigma=2$ yields the best results across all sampling rates.

In the pursuit of identifying the optimal Fourier modes, we tested different values, finding that 20 Fourier modes achieved the best performance. Considering the model architecture design, we trained models with varying numbers of layers. Notably, the configuration with four Fourier blocks and GELU activation function demonstrated superior performance.

Overall, our ablation studies confirmed the relative optimality of our pipeline regarding hyperparameter changes, emphasizing the significance of incorporating normal field information and utilizing specific parameter settings for the smoothing Gaussian and Fourier modes. The chosen architecture with four Fourier blocks and GELU activation function exhibited the most favorable performance, contributing to the effectiveness of our approach.

    \section{Conclusion}
This paper presents \ours{}, a groundbreaking approach to 3D shape reconstruction that significantly outperforms state-of-the-art techniques, particularly in low-sampling scenarios with point clouds ranging from $3\,000$ to $25\,000$ points. Even in high-sampling domains, \ours{} competes favorably with other methods, demonstrating superior capability in generating nearly precise reconstructions with a substantial data set, encompassing around $250\,000$ points.

The core strength of \ours{} originates from its sophisticated deep neural network architecture that seamlessly integrates Fourier layers, recognized as Fourier Neural Operators. These layers adeptly address the Poisson equation, enabling the generation of a voxel grid, which serves as the 3D reconstruction output. The strategic implementation of Otsu's thresholding on the reconstruction, coupled with the utilization of marching cubes to extract a mesh representation, underscores the robustness and versatility inherent in \ours{}.

One of the standout advantages of \ours{} is its one-shot super-resolution capabilities inherited from Fourier Neural Operators. This innovative feature facilitates training on low-resolution voxel grids ($64{\times}64{\times}64$) while ensuring comparable performance during evaluation on higher-resolution grids ($128{\times}128{\times}128$). This is a marked advancement over traditional convolutional neural networks, where errors tend to escalate exponentially with the augmentation of the input dimension.

The contributions of this paper extend beyond 3D shape reconstruction, encouraging further investigation of Fourier Neural Operators across various computer vision tasks. The observed resolution agnosticism, though challenged by the need for a predetermined resolution of training data, presents an avenue for additional exploration. The requirement to specify the resolution of training data in advance often leads to a reduction of all training data to a set resolution, resulting in the loss of fine details within images. This is a significant challenge shared by similar methods aimed at solving partial differential equations. However, with sufficient computational resources, training at a higher resolution using more capable GPUs becomes a feasible option.

This work alludes to the potential of delving deeper into optimizing \ours{} to enhance its performance and generalization further. The prospect of exploring alternative architectures, incorporating additional regularization techniques, or investigating different loss functions appears promising in tackling specific challenges inherent in shape reconstruction. Particularly captivating is the notion of extending our approach beyond the confinement of regular grid representations by employing graph neural operators to manage highly detailed meshes.

In summary, the results presented in this thesis not only provide a robust methodology for 3D shape reconstruction but also lay the foundation for a broader application of Fourier Neural Operators in the broad field of computer vision. The resolution agnosticism demonstrated here, coupled with the suggestion of further optimization of \ours{} or exploration of alternative architectures, suggests a positive path forward for advancing shape reconstruction and other areas of computer vision.

    \section{Acknowledgments}

The work of H. Andrade-Loarca was supported by the German Research Foundation under
Grant DFG-SFB/TR 109, Project C09. J. Hege was supported by in part by the Konrad Zuse
School of Excellence in Reliable AI (DAAD) as well as the German Research Foundation under
Grant DFG-SFB/TR 109, Project C09. This work of D. Cremers was supported by the ERC Advanced Grant SIMULACRON. G. Kutyniok acknowledges support by the Konrad Zuse
School of Excellence in Reliable AI (DAAD), the Munich Center for
Machine Learning (BMBF) as well as the German Research Foundation under
Grant DFG-SFB/TR 109, Project C09. 
    {
        \small
        \bibliographystyle{ieeenat_fullname}
        \bibliography{references}
    }
    
    % WARNING: do not forget to delete the supplementary pages from your submission 
    % \input{sec/X_suppl}
\end{document}